\documentclass[10pt,conference]{IEEEtran}

\usepackage{booktabs} % For formal tables
\usepackage{multirow}
\usepackage[position=bottom]{subfig}
\usepackage{multicol}
\usepackage{cite}
\usepackage{flushend}
\usepackage{booktabs}
\usepackage[pdftex]{graphicx}
\usepackage{lipsum}

\usepackage[cmex10]{amsmath}
\newcommand{\eg}{{\textit{e.g.}}}
\newcommand{\ie}{{\textit{i.e.}}}
\newcommand{\etal}{{\textit{et al.}}}

\begin{document}
\title{Investigating Recurrent Neural Network Memory Structures using Neuro-Evolution}
\author{
\IEEEauthorblockN{
Alexander Ororbia*\IEEEauthorrefmark{1},
AbdElRahman ElSaid\IEEEauthorrefmark{2},
Travis Desell*\IEEEauthorrefmark{2}
}

\thanks{* Both authors contributed equally.}

\IEEEauthorblockA{
Department of Computer Science\IEEEauthorrefmark{1}, Department of Software Engineering\IEEEauthorrefmark{2}\\
Rochester Institute of Technology\\
20 Lomb Memorial Dr.\\
Rochester, New York 14623 \\
Email: ago@cs.rit.edu, aae8800@rit.edu, tjdvse@rit.edu
}
}

\maketitle

\begin{abstract}
    This paper presents a new algorithm, Evolutionary eXploration of Augmenting Memory Models (EXAMM), which is capable of evolving recurrent neural networks (RNNs) using a wide variety of memory structures, such as $\Delta$-RNN, GRU, LSTM, MGU and UGRNN cells.  EXAMM evolved RNNs to perform prediction of large-scale, real world time series data from the aviation and power industries. These data sets consist of very long time series (thousands of readings), each with a large number of potentially correlated and dependent parameters. Four different parameters were selected for prediction and EXAMM runs were performed using each memory cell type alone, each cell type and simple neurons, and with all possible memory cell types and simple neurons. Evolved RNN performance was measured using repeated $k$-fold cross validation, resulting in $2420$ EXAMM runs which evolved $4,840,000$ RNNs in ${\sim}24,200$ CPU hours on a high performance computing cluster. Generalization of the evolved RNNs was examined statistically, providing interesting findings that can help refine the RNN memory cell design as well as inform future neuro-evolution algorithms development. 

%EXAMM source code along with the data sets utilized have been made publicly available to facilitate reproducibility and further research in time series data prediction as to the authors knowledge, there are no other data sets of this type and scale publicly available.

\end{abstract}


\section{Introduction}
\label{sec:introduction}

This work is motivated by a major open question in the field of artificial neural network (ANN) research: \emph{What neural memory structures appear to be optimal for time-series prediction?} Conducting such a search over ANN structures entails manual, primarily human-driven labor and activity. As more advances are made in the field, the number of possible architecture variations and modifications explodes combinatorially. The growing space of architecture structures combined with the limited, often simple heuristic local search that can be conducted by human experts means that efficiently finding architectures that generalize well while still maintaining low parameter complexity (given that regularization is important for most data sample sizes) makes for a nearly impossible, largely intractable search problem.

In natural biological systems, the process of evolution, over long time-spans, endows organisms with various inductive biases that allow them to adapt and learn their environment quickly and readily. It is thought that these inductive biases are what provide infants the ability to quickly learn complex pattern recognition/detection functions with limited data across various sensory modalities~\cite{morton1991conspec,fantz1961origin}, such as in visual and speech sensory domains. While artificial forms of evolution, such as the classical genetic algorithm \cite{holland1992genetic}, are significantly simplified from the actual evolutionary process that creates useful inductive biases to drive development and survival of organisms at large, these optimization procedures offer the chance to develop non-human centered ways of generating useful and even potentially optimal neural architectures.

While neuro-evolution (applying evolutionary processes to the development of ANNs) been used in searching the space of feed forward and even convolutional architectures for tasks involving static inputs \cite{gomez2008accelerated,salama2014novel,zoph2016neural,xie2017geneticcnn,suganuma2017genetic,sun-arxiv-evocnn-2017,mikkulainen2017codeepneat,real2017evolution,stanley2002evolving,stanley2009hypercube}, less effort has been put into exploring the evolution of recurrent memory structures that operate with complex time based data sequences and uncovering what forms and structures the neuro-evolutionary process finds. Insights extracted from the evolved architectures can serve as inductive biases for subsequent research in neural network design and development.

The evolution of recurrent neural networks (RNNs) poses significant challenges above beyond the already challenging task of evolving feed forward and convolutional neural networks. First, RNNs are more challenging to train due to issues with exploding and vanishing gradients which occur when unrolling a RNN over a long time series with the backpropagation through time (BPTT) algorithm~\cite{werbos1990backpropagation}. Due to this issue, development of recurrent memory cells which can preserve memory and long term dependencies while alleviating the exploding and vanishing gradient problem has been an area of significant study~\cite{hochreiter1997long,chung2014empirical,collins2016capacity,zhou2016minimal,ororbia2017diff}. Further, as in the case of time-varying data, input samples are strictly ordered in time and thus induce long-term dependencies that any useful stateful adaptive process/model must extract in order to generalize. The complexity of time-series tasks varies, entailing the prediction of a particular variable of interest over time or even constructing a full generative model of all the available variables, perhaps additionally involving other non-trivial tasks such as data imputation.  

Due to these issues, RNNs for time series data prediction can require the use of a variety of memory cell structures in addition to recurrent connections spanning different time spans. To optimize within this large search space, Evolutionary eXploration of Augmenting Memory Models (EXAMM) integrates an extensible collection of different complex memory cell types with simple neuronal building blocks and recurrent connections of varying time spans. EXAMM was used to evolve RNNs with various cell types to predict time series data from two real-world, large-scale data sets. 

In total, 2420 EXAMM runs were done using repeated $k$-fold cross valdation, generating a large set of evolved RNNs to analyze statistically. Results are particularly interesting in that while they show allowing selection from any of the available memory cell structures can provide very well performing networks in many test cases; it does come at the cost of reliability (how well the evolved RNNs perform in an average case).  Further, even small modifications to the neuro-evolutionary process, such as allowing simple neurons, can generally improve predictions, but they can also have unintended consequences -- in some examples it meant the difference between a memory cell structure performing the best as opposed to the worst. The authors hope that these results can help guide not only the further development of neuro-evolution algorithms, but also help refine and inform the development of new memory cell structures and human designed RNNs.

%Given the high complexity of conducting a primarily human-centered space, developing an efficient, effective neuro-evolutionary metaheuristic would be invaluable in the face of ever-increasing datasets. Beyond the development of a powerful metaheuristic, we seek to extract any useful insights found in the solutions yielded by our algorithm, especially with respect to the time series datasets explored in this paper.

%Neuro-evolution approaches typically fall under one of two broad classes--standard, where synaptic weights are evolved in parallel to the ANN structure, and memetic, which combine an evolutionary/population-based search with a local improvement procedure for individual solutions/agents. Our method falls under the former category since we do not include a local search procedure, such as stochastic hill-climbing, though such as an extension would readily work within the EXAMM framework.

%In this paper, we will develop a rather general neuro-evolutionary algorithm, 

\section{Evolving Recurrent Networks}
\label{sec:related_work}

Several methods for evolving NN topologies along with weights have been searched and deployed, with NeuroEvolution of Augmenting Topologies (NEAT)~\cite{stanley2002evolving} perhaps being the most well-known. EXAMM differs from NEAT in that it includes more advanced node-level mutations (see Section~\ref{sec:node_mutations}), and utilizes Lamarckian weight initialization (see Section~\ref{sec:lamarckian_weight_initialization}) along with backpropagation to evolve the weights as opposed to a simpler less efficient genetic strategy. Additionally, it has been developed with large-scale concurrency in mind, and utilizes an asynchronous steady state approach which has been shown to facilitate scalability to potentially millions of compute nodes~\cite{desell-phd-2009}.

%It is a genetic algorithm that evolves increasingly complex neural network topologies, while at the same time evolving the connection weights. Genes are tracked using historical markings with innovation numbers to perform crossover among different structures and enable efficient recombination. Innovation is protected through speciation and the population initially starts small without hidden layers and gradually grows through generations~\cite{annunziato2002adaptive,larochelle2009exploring,kandel2000principles}. 

Other recent work by Rawal and Miikkulainen has investigated an information maximization objective~\cite{rawal2016evolving} strategy to evolve RNNs. This strategy essentially utilizes NEAT with Long Short Term Memory (LSTM) neurons instead of regular neurons. EXAMM provides a more in-depth study of the performance of possible recurrent cell types as it examines both simple neurons and four other cell structures beyond LSTMs. Rawal and Miikulainen have also utilized tree-based encoding~\cite{rawal-evolving-rnns-2018} to evolve recurrent cellular structures within fixed architectures built of layers of the evolved cell types. Combining this evolution of cell structure along with the overall RNN structure stands as interesting future work.

Ant colony optimization (ACO) has also been investigated as a way to select which connections should be utilized in RNNs and LSTM RNNs by Desell and ElSaid~\cite{desell2015evolving,elsaid2018optimizing,elsaid2018using}. In particular, this ACO approach was shown to reduce the number of trainable connections in half while providing a significant improvement in predictions of engine vibration~\cite{elsaid2018optimizing}. However, this approach works within a fixed RNN architecture and cannot evolve an overall RNN structure.

\section{Evolutionary eXploration of Augmenting Memory Models}
\label{sec:examm}

The EXAMM algorithm presented in this work expands on an earlier algorithm, EXALT~\cite{desell-evostar-2019} which can evolve RNNs with either simple neurons or LSTM cells. It further refines EXALT's mutation operations to reduce hyperparameters using statistical information from parental RNN genomes. Also, EXALT only used a single steady state population, and EXAMM expands on this to use islands, which have been shown by Alba and Tomassini to greatly improve performance of distributed evolutionary algorithms, potentially providing superlinear speedup~\cite{alba2002parallelism}. A master process maintains each population of islands, and generates new RNNs form islands in a round robin manner to be trained upon request by workers. When a worker completes training a RNN, it is inserted into the island it was generated from if its fitness (mean squared error on the test data) is better than the worst in the island, and then the worst RNN in the island is removed. This asynchrony is particularly important as the generated RNNs will have different architectures, each taking a different amount of time to train. By having a master process control the population, workers can complete the training of the generated RNNs at whatever speed they can and the algorithm is naturally load balanced. This allows EXAMM to scale to the number of available processors, having a population size independent of processor availability, unlike synchronous parallel evolutionary strategies. The EXAMM codebase has a multithreaded implementation for multicore CPUs as well as an MPI~\cite{mpif94mpi} implementation for use on high performance computing resources.

\subsection{Mutation and Recombination Operations}
\label{sec:mutation}

\begin{figure}
    \centering
    \subfloat[The edge between Input 1 and Output 1 is selected to be split. A new node with innovation number (IN) 1 is created. ]{
        \includegraphics[width=0.45\textwidth]{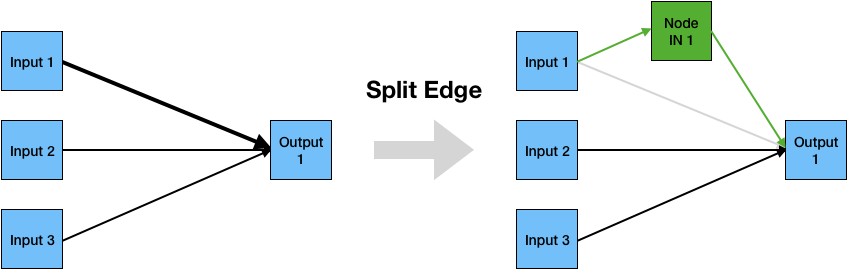}
    }\hfill
    \subfloat[Input 3 and Node IN 1 are selected to have an edge between them added.]{
        \includegraphics[width=0.45\textwidth]{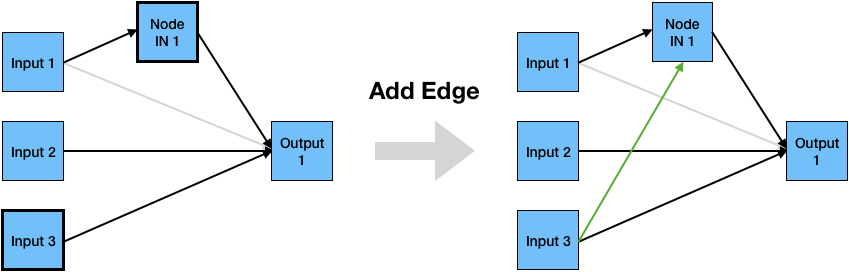}
    }

    \subfloat[The edge between Input 3 and Output 1 is enabled.]{
        \includegraphics[width=0.45\textwidth]{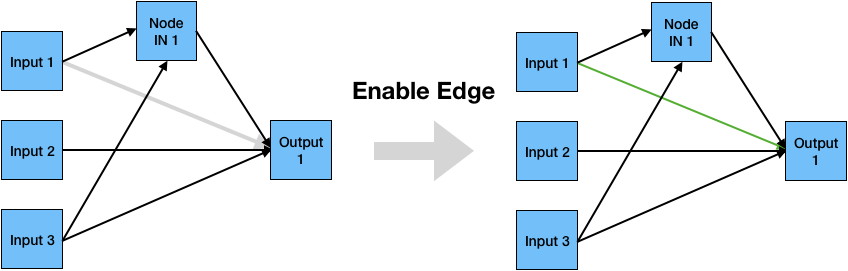}
    }\hfill
    \subfloat[A recurrent edge is added between Output 1 and Node IN 1]{
        \includegraphics[width=0.45\textwidth]{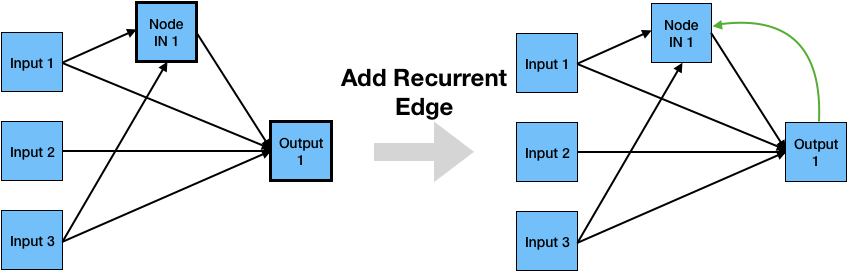}
    }

    \subfloat[The edge between Input 3 and Output 1 is disabled.]{
        \includegraphics[width=0.45\textwidth]{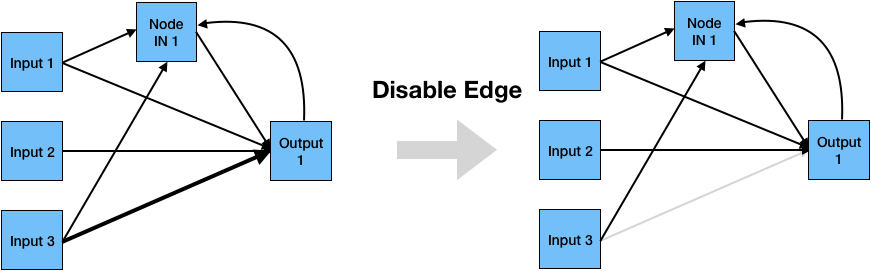}
    }
    \caption{\label{fig:edge_mutations} Edge mutation operations.}
\end{figure}

RNNs are evolved with edge-level operations, as done in NEAT, as well as with new high level node mutations as in EXALT and EXACT. Whereas NEAT only requires innovation numbers for new edges, EXAMM requires innovation numbers for both new nodes, new edges and new recurrent edges. The master process keeps track of all node, edge and recurrent edge innovations made, which are required to perform the crossover operation in linear time without a graph matching algorithm.  Figures~\ref{fig:edge_mutations} and~\ref{fig:node_mutations_1} display a visual walkthrough of all the mutation operations used by EXAMM. Nodes and edges selected to be modified are highlighted, and then new elements to the RNN are shown in green. Edge innovation numbers are not shown for clarity. Enabled edges are in black, disabled edges are in grey.

It should be noted that for the following operations, whenever an edge is added, unless otherwise specified, it is probabilistically selected to be a recurrent connection with the following recurrent probability: $p = \frac{n_{re}}{n_{ff} + n_{re}}$, where $n_{re}$ is the number of enabled recurrent edges and $n_{ff}$ is the number of enabled feed forward edges in the parent RNN.  A recurrent connection will go back a randomly selected number of time steps with bound specified as search parameters (in this work, 1 to 10 time steps), allowing for recurrent connections of varying time spans.  Any newly created node is selected uniformly at random as a simple neuron or from the memory cell types specified by the EXAMM run input parameters.

\subsubsection{Edge Mutations:}
\label{sec:edge_mutations}

\paragraph{Disable Edge} This operation randomly selects an enabled edge or recurrent edge in a RNN genome and disables it so that it is not used.  The edge remains in the genome. As the \emph{disable edge} operation can potentially make an output node unreachable, after all mutation operations have been performed to generate a child RNN genome, if any output node is unreachable that RNN genome is discarded without training.

\paragraph{Enable Edge} If there are any disabled edges or recurrent edges in the RNN genome, this operation selects a disabled edge or recurrent edge at random and enables it.

\paragraph{Split Edge} This operation selects an enabled edge at random and disables it. It creates a new node and two new edges, and connects the input node of the split edge to the new node, and the new node to the output node of the split edge. If the split edge was recurrent, the new edges will also be recurrent (with the same time skip); otherwise they will be feed forward.

\paragraph{Add Edge} This operation selects two nodes $n_1$ and $n_2$ within the RNN Genome at random, such that $depth_{n_1} < depth_{n_2}$ and such that there is not already an edge between those nodes in this RNN Genome. Then it adds an edge from $n_1$ to $n_2$. 

\paragraph{Add Recurrent Edge} This operation selects two nodes $n_1$ and $n_2$ within the RNN Genome at random and then adds a recurrent edge from $n_1$ to $n_2$, selecting a time span as described before. The same two nodes can be connected with multiple recurrent connections, each spanning different times; however it will not create a duplicate recurrent connection with the same time span.

\subsubsection{Node Mutations:}
\label{sec:node_mutations}
    
\begin{figure}
    \centering
    \subfloat[A node with IN 2 is selected to be added at a depth between the inputs \& Node IN 1. Edges are randomly added to Input 2 and 3, and Node IN 1 and Output 1.]{
        \includegraphics[width=0.45\textwidth]{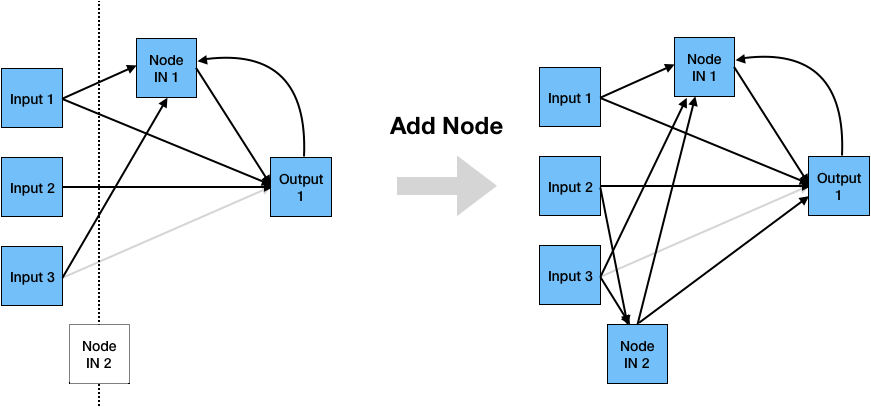}
    }

    \subfloat[Node IN 1 is selected to be split. It is disabled with its input/output edges. It is split into Nodes IN 3 and 4, which get half the inputs. Both have an output edge to Output 1 since there was only one output from Node IN 1.]{
        \includegraphics[width=0.45\textwidth]{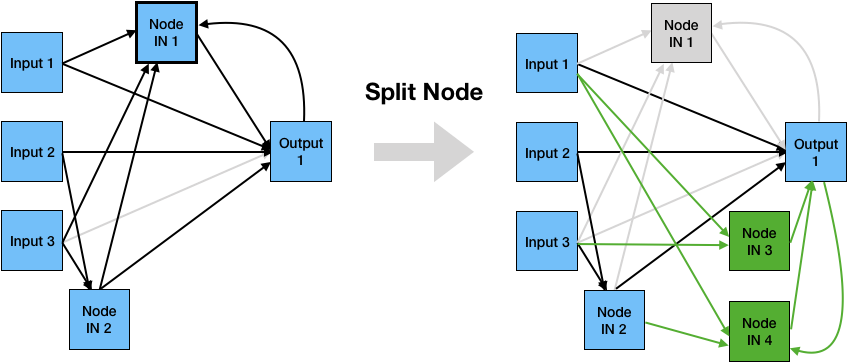} 
    }

    \subfloat[Node IN 2 and 3 are selected for a merger (input/output edges are disabled). Node IN 5 is created with edges between all their inputs/outputs.]{
        \includegraphics[width=0.45\textwidth]{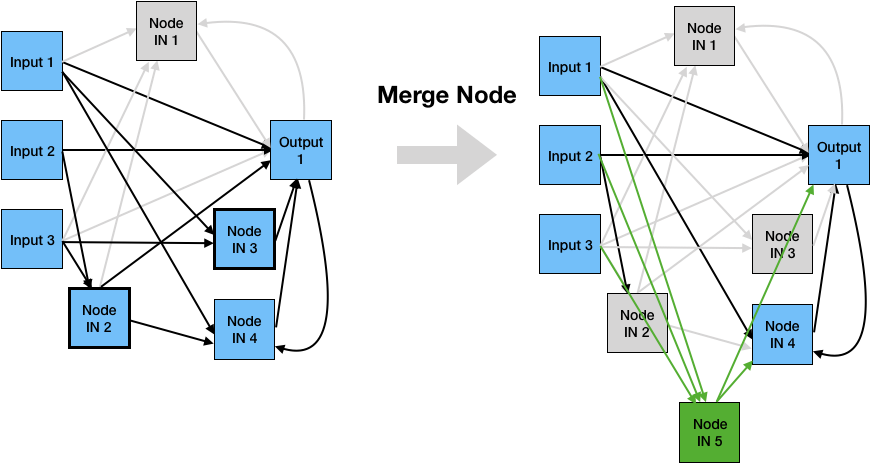}
    }
    \caption{\label{fig:node_mutations_1} Node mutation operations.}
\end{figure}

\begin{figure}
    \ContinuedFloat
    \centering
    \subfloat[Node IN 1 is selected to be enabled, along with all its input and output edges.]{
        \includegraphics[width=0.45\textwidth]{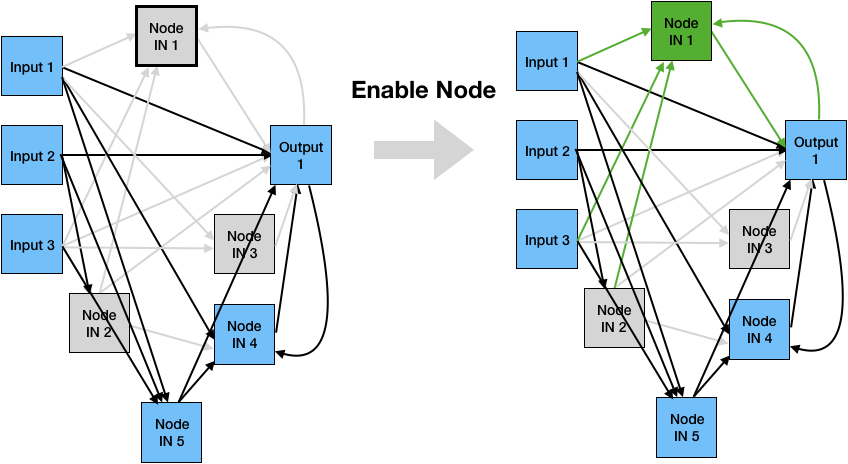}
    }

    \subfloat[Node IN 5 is selected to be disabled, along with all its input and output edges.]{
        \includegraphics[width=0.45\textwidth]{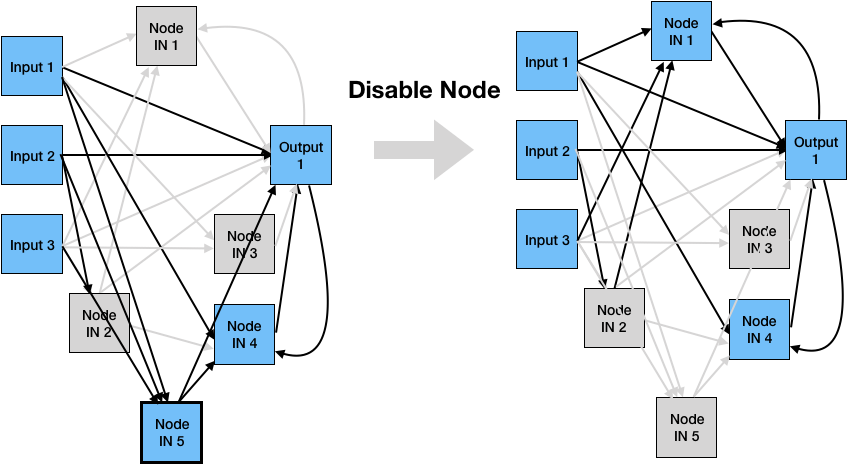}
    }

    \caption{\label{fig:node_mutations_2} Node mutation operations (continued).}
\end{figure}

\paragraph{Disable Node} This operation selects a random non-output node and disables it along with all of its incoming and outgoing edges. Note that this allows for input nodes to be dropped out, which can be useful when it is not previously known which input parameters are correlated to the output.

\paragraph{Enable Node} This operation selects a random disabled node and enables it along with all of its incoming and outgoing edges.

\paragraph{Add Node} This operation selects a random depth between 0 and 1, noninclusive. Given that the input node is always depth 0 and the output nodes are always depth 1, this depth will split the RNN in two. A new node is created at that depth, and the number of input and output edges and recurrent edges are generated using normal distributions with mean and variances equal to the mean and variances for the of input/output edges and recurrent edges of all nodes in the parent RNN.  

\paragraph{Split Node} This operation takes one non-input, non-output node at random and splits it. This node is disabled (as in the disable node operation) and two new nodes are created at the same depth as their parent. At least one input and one output edge are assigned to each of the new nodes, of a duplicate type from the parent, with the others being assigned randomly, ensuring that the newly created nodes have both inputs and outputs. 

\paragraph{Merge Node} This operation takes two non-input, non-output nodes at random and combines them.  Selected nodes are disabled (as in the disable node operation) and a new node is created at depth equal to the average of its parents. This node is connected to the inputs and outputs of its parents with a duplicate type from the parent; as input edges connected to lower depth nodes and output edges connect to greater depth nodes. 

\subsubsection{Other Operations:}

%\paragraph{Crossover} utilizes two hyperparameters, the \emph{more fit crossover rate} and the \emph{less fit crossover rate}. Two parent RNN genomes are selected, and the child RNN genome is generated from every edge that appears in both parents. Edges that only appear in the more fit parent are added randomly at the \emph{more fit crossover rate}, and edges that only appear in the less fit parent are added randomly at the \emph{less fit crossover rate}. Edges not added by either parent are also carried over into the child RNN genome, however they are set to disabled. Nodes are then added for each input and output of an edge.  If the more fit parent has a node with the same innovation number, it is added from the more fit parent. 

\paragraph{Crossover} creates a child RNN using all reachable nodes and edges from two parents. A node or edge is reachable if there is a path of enabled nodes and edges from an input node to it as well as a path of enabled nodes and edges from it to an output node, \ie, a node or edge is reachable if it actually affects the RNN. Crossover can be done either within an island (\emph{intra-island}) or between islands (\emph{inter-island}). Inter-island crossover selects a random parent in the target island, and the best RNN from the other islands.

\paragraph{Clone} creates a copy of the parent genome, initialized to the same weights.  This allows a particular genome to continue training in cases where further training may be more beneficial than performing a mutation or crossover.

\subsection{Lamarckian Weight Initialization}
\label{sec:lamarckian_weight_initialization}

For RNNs generated during population initialization, the weights are initialized uniformly at random between $-0.5$ and $0.5$. Biases and weights for new nodes and edges are initialized from a normal distribution based on the average, $mu$, and variance, $\sigma^2$, of the parents' weights. However, RNNs generated through mutation or crossover re-use parental weights, allowing the RNNs to train from where the parents are left off, \ie, \emph{``Lamarckian'' weight initialization}.

During crossover, in the case where an edge or node exists in both parents, the child weights are generated by recombining the parents' weights.  Given a random number $-0.5 <= r <= 1.5$, a child's weight $w_c$ is set to $w_c = r(w_{p2} - w_{p1}) + w_{p1}$, where $w_{p1}$ is the weight from the more fit parent, and $w_{p2}$ is the weight from the less fit parent. This allows the child weights to be set along a gradient calculated from the weights of the two parents.

This weight initialization strategy is particularly important as newly generated RNNs do not need to be completely retrained from scratch. In fact, the RNNs only need to be trained for a few epochs to investigate the benefits of newly added structures.  In this work, the generated RNNs are only trained for 10 epochs (see Section~\ref{sec:experimental_design}), where training a static RNN structure from scratch may require hundreds or even thousands of epochs.

\subsection{A Collection of Memory Cells}
\label{sec:mem_models}
Node types can range quite a bit in terms of complexity and their design largely governs the form of the underlying memory structure. Simple neurons can be evolved into generalized versions of traditional Elman and Jordan neurons as EXAMM adds recurent connections. Below describes how simple neurons can evolve to generalized Elman and Jordan neurons, as well as complex cells such as the Delta-RNN, the minimally gated unit (MGU) \cite{zhou2016minimal}, the update-gated RNN (UGRNN) \cite{collins2016capacity}, the Gated Recurrent Unit (GRU) \cite{chung2014empirical}, and Long Short Term Memory (LSTM) \cite{hochreiter1997long}. %These cell models together make up the internal library accessible to EXAMM for possible selection during its metaheuristic search. 

\noindent
\textbf{Elman, Jordan and Arbitrary Recurrent Connections:} 
In EXAMM, simple neurons are represented as potential recurent neurons with both recurrent and feed forward inputs. $I$ is the set of all nodes with a feed forward connection to simple neuron $j$ while $R$ is the set of all nodes with a recurrent connection to simple neuron $j$. At time step $t$, the output is a weighted summation of all feed forward inputs, where $w_{ij}$ is the feed forward weight connecting node $i$ to node $j$, plus a weighted summation of recurrent inputs, where $v_{rjk}$ is the recurrent weight from node $r$ at time step $t-k$ to the node $j$, where $k$ is the time span of the recurrent connection. Thus the state function $s$ for a computing a simple neuron is:\footnote{The bias is omitted for clarity and simplicity of presentation.}
\vspace{-2mm}
\begin{align*}
    s_j(t) &= \phi_s \bigg( \sum_{i \in I} w_{ij} \cdot s_i(t) + \sum_{r \in R,k} v_{rjk} \cdot s_r(t-k) \bigg)
\vspace{-8mm}
\end{align*}
The overall state is a linear combination of the projected input and an affine transformation of the vector summary of the past. The post-activation function, $\phi_s(\cdot)$, can be any differentiable element-wise function, however for the purposes of this work it was limited to the hyperbolic tangent $\phi(v) = tanh(v) = (e^{(2v)} - 1) / (e^{(2v)} + 1)$. %Future work will involve selection of a set of activation functions including the logistic sigmoid and linear rectifier.

%The post-activation function $\phi_s(\cdot)$ can be any differentiable element-wise function, such as the logistic sigmoid $\phi_s(v) = \sigma(v) = 1/(1 + e^{-v})$, the hyperbolic tangent $\phi(v) = tanh(v) = (e^{(2v)} - 1) / (e^{(2v)} + 1)$, or the linear rectifier $\phi_s(v) = relu(v) = max(0, v)$. Note that we recover the pure feedforward node by simply setting and fixing $v_{ij} = 0$, which is also one possible node that EXAMM may select during the search process.

%EXAMM also evolves a form of recurrent node inspired by the Jordan RNN \cite{jordan1997serial}, which, in the original model, uses output-to-state recurrent connections instead of state-to-state connections connections to compute its internal state. In EXAMM, this is generalized by allowing nodes to feature cross-layer recurrent connections, \ie, the node is not restricted to only be connected itself but also to connect to a node at any other depth. Recurrent connections in Jordan/Elman RNNs traditionally only span one time step, however EXAMM generalizes this as well to allow connections spanning time steps between user-specified markers (in this case 1 \& 10).

Elman-RNNs~\cite{elman1990finding} are the simplest of all RNNs, where recurrent nodes connections to themselves and potentailly all other hidden nodes in the same layer. Jordan-RNNs \cite{jordan1997serial} have recurrent connections from output node(s) to hidden node(s). EXAMM can evolve Elman connections when the \emph{add recurrent edge} mutation adds a recurrent edge from a simple neuron back to itself, and Jordan connections when it adds a recurrent edge from an output to a simple neuron. It generalizes these structures by allowing varying time spans (Jordan and Elmann network traditionally span only one time step) and arbitrary recurrent connections between from any pair of simple neurons or cells, i.e., cross-layer recurrent connections.

\noindent
\textbf{The Delta RNN ($\Delta$-RNN) Cell:} 
For models more complex than the simple RNN model, we looked to derive a vast set of gated neural architectures unified under a recent framework known as the Differential State Framework (DSF) \cite{ororbia2017diff}. A DSF neural model is essentially a composition of an inner function used to compute state proposals and an outer ``mixing'' function used to decide how much of a state proposal is incorporated into a slower moving state, \ie, the longer term memory. These models are better equipped to handle the vanishing gradient problem induced by the difficulty of learning long-term dependencies over sequences \cite{bengio1994learning}. Other models that can be derived under the DSF include the LSTM, the GRU, and the MGU \cite{ororbia2017diff}. One of the simplest DSF models is the $\Delta$-RNN, which has been shown to perform competitively with more complex memory models in problems ranging from language modeling \cite{ororbia2017diff,ororbia2018using} to image decoding \cite{ororbia2018learned}. With $\{\alpha,\beta_1,\beta_2,b_j,m\}$ as learnable coefficient scalars, the $\Delta$-RNN state is defined as:
%\vspace{-2mm}
\begin{align*}
e^{w}_j &= \sum_{i \in I} w_{ij} \cdot s_i(t) + \sum_{r \in R,k} v_{rjk} \cdot s_r(t-k),\quad e^{v}_j = m \cdot s_j(t-1) \\
d^1_j &= \alpha \cdot e^{v}_j \cdot e^{w}_j,\quad d^2_j = \beta_1 \cdot e^{v}_j + \beta_2 \cdot e^{w}_j, \quad r_j = \sigma(e^{w}_j + b_j) \\
\widetilde{s}_j(t) &= \phi_s(d^1_j + d^2_j), \quad s_j(t) = \Phi_s( (1 - r_j) \cdot \widetilde{s}_j(t) + r_j \cdot s_j(t-1) \\
\vspace{-10mm}
\end{align*}

\noindent
\textbf{The Long-Short Term Memory (LSTM):} The LSTM \cite{hochreiter1997long} is one of the most commonly used gated neural model when modeling sequential data. The original motivation behind the LSTM was to implement the ``constant error carousal'' in order to mitigate the problem of vanishing gradients. This means that long-term memory can be explicitly represented with a separate cell state $\mathbf{c}_t$. 

The LSTM state function (without extensions, such as ``peephole'' connections) is implemented as follows:
\begin{align*}
f_j &= \sigma(\sum_{i \in I} w^f_{ij} \cdot s_i(t) + \sum_{r \in R,k} v^f_{rjk} \cdot s_r(t-k)) \\
i_j &= \sigma(\sum_{i \in I} w^i_{ij} \cdot s_i(t) + \sum_{r \in R,k} v^i_{rjk} \cdot s_r(t-k)) \\
\widetilde{c}_j &= tanh(\sum_{i \in I} w^c_{ij} \cdot s_i(t) + \sum_{r \in R,k} v^c_{ijk} \cdot s_r(t-k)) \\
o_j &= \sigma(\sum_{i \in I} w^o_{ij} \cdot s_i(t) + \sum_{r \in R,k} v^o_{ijk} \cdot s_r(t-k)) \\
c_j(t) &= f_j(t) \cdot c_j(t-1) + i_j \cdot \widetilde{c}_j, \quad s_j(t) = o_j \cdot \phi_s( c_j(t) )
\end{align*}
where we depict the sharing of the transformation function's output across the forget ($\mathbf{f}_t$), input ($\mathbf{i}_t$), cell-state proposal ($\widetilde{\mathbf{c}}_t$), and output ($\mathbf{o}_t$) gates.
The LSTM is by far the most parameter-hungry of the DSF models we explore.

\noindent
\textbf{The Gated Recurrent Unit (GRU):}
The Gated Recurrent Unit (GRU; \cite{chung2014empirical}) can be viewed as an early attempt to simplify the LSTM. Among the changes made, the model fuses the LSTM input and forgets gates into a single gate, and merges the cell state and hidden state back together.
The state function based on the GRU is calculated using the following equations:
\begin{align}
z_j &= \sigma(\sum_{i \in I} w^z_{ij} \cdot s_i(t) + \sum_{r \in R,k} v^z_{ijk} \cdot s_r(t-k)) \\
r_j &= \sigma(\sum_{i \in I} w^r_{ij} \cdot s_i(t) + \sum_{r \in R,k} v^r_{ijk} \cdot s_r(t-k)) \\
\widetilde{s}_j(t) &= \phi_s(\sum_{i \in I} w^s_{ij} \cdot s_i(t) + \sum_{r \in R,k} v^s_{ijk} \cdot (r_j \cdot s_r(t-k)) \\
s_j(t) &= z_j \cdot \widetilde{s}_j + (1 - z_j) \cdot s_j(t-1)
\label{state:gru}
\end{align}
noting that $\phi_s(v) = tanh(v)$.

\noindent
\textbf{The Minimally-Gated Unit (MGU):} The MGU model is very similar in structure to the GRU, reducing number of required parameters by merging its reset and update gates into a single forget gate \cite{zhou2016minimal}. The state computation proceeds as follows:
\begin{align}
f_j &= \sigma(\sum_{i \in I} w^f_{ij} \cdot s_i(t) + \sum_{r \in R,k} v^f_{ijk} \cdot s_r(t-k)) \\
\widetilde{s}_j(t) &= \phi_s(\sum_{i \in I} w^s_{ij} \cdot s_i(t) + \sum_{r \in R,k} v^s_{ijk} \cdot (f_j \cdot s_r(t-k)) \\
s_j(t) &= z_j \cdot \widetilde{s}_j + (1 - z_j) \cdot s_j(t-1)\mbox{.}
\label{state:mgu}
\end{align}

\noindent
\textbf{The Update-Gated RNN (UGRNN):} The UGRNN \cite{collins2016capacity} updates are defined in the following manner:
\begin{align}
c_j &= \phi_s(\sum_{i \in I} w^c_{ij} \cdot s_i(t) + \sum_{r \in R,k} v^c_{ijk} \cdot s_r(t-k)) \\
g_j &= \sigma(\sum_{i \in I} w^g_{ij} \cdot s_i(t) + \sum_{r \in R,k} v^g_{ijk} \cdot s_r(t-k)) \\
s_j(t) &= g_j \cdot s_j(t-1) + (1 - g_j) \cdot c_j  \mbox{.}
\label{state:ugrnn}
\end{align}
The UGRNN, though more expensive than the $\Delta$-RNN, is a simple model, essentially working like an Elman-RNN with a single update gate. This extra gate decides whether a hidden state is carried over from the previous time step or if the state should be updated.

%Since EXAMM operates with a focus on single unit processing elements we will present the formulation of each unit type from the same perspective (instead of the usual vectorized forms which deals with collections or layers of neuronal elements, as is done in other work).
\section{Open Data and Reproducibility}
\label{sec:data_sets}

This work utilizes two data sets to benchmark the memory cells and RNNs evolved by EXAM. The first comes from a selection of 10 flights worth of data from the National General Aviation Flight Information Database (NGAFID)~\cite{ngafid} and the other comes from a coal-fired power plant (which has requested to remain anonymous). Both datasets are multivariate (26 and 12 parameters, respectively), non-seasonal, and the parameter recordings are not independent.  Furthermore, they are very long -- the aviation time series range from 1 to 3 hours worth of per-second data while the power plant data consists of 10 days worth of per-minute readings. These data sets are provided openly through the EXAMM GitHub repository\footnote{https://github.com/travisdesell/exact}, in part for reproducibility, but also to provide a valuable resource to the field. To the authors' knowledge, real world time series data sets of this size and at this scale are not freely available.

{\it RPM} (rotations per minute) and {\it pitch} were selected as prediction parameters from the aviation data since RPM is a product of engine activity, with other engine-related parameters being correlated, and since pitch is directly influenced by pilot controls, making it particularly challenging to predict. For the coal plant data, \emph{main flame intensity} and \emph{supplementary fuel flow} were selected as parameters of interest. Similar to the choices from the NGAFID data, main flame intensityy is mostly a product of conditions within the (coal) burner, while supplementary fuel flow is more directly controlled by human operators.

\subsection{Aviation Flight Recorder Data}

With permission, data from 10 flights was extracted from the NGAFID. Each of the 10 flight data files last over an hour, and consist of per-second data recordings from 26 parameters:

%\begin{multicols}{2}
\begin{enumerate}
\item Altitude Above Ground Level (AltAGL)
\item Engine 1 Cylinder Head Temperature 1 (E1 CHT1)
\item Engine 1 Cylinder Head Temperature 2 (E1 CHT2)
\item Engine 1 Cylinder Head Temperature 3 (E1 CHT3)
\item Engine 1 Cylinder Head Temperature 4 (E1 CHT4)
\item Engine 1 Exhaust Gas Temperature 1 (E1 EGT1)
\item Engine 1 Exhaust Gas Temperature 2 (E1 EGT2)
\item Engine 1 Exhaust Gas Temperature 3 (E1 EGT3)
\item Engine 1 Exhaust Gas Temperature 4 (E1 EGT4)
\item Engine 1 Oil Pressure (E1 OilP)
\item Engine 1 Oil Temperature (E1 OilT)
\item Engine 1 Rotations Per minute (E1 RPM)
\item Fuel Quantity Left (FQtyL)
\item Fuel Quantity Right (FQtyR)
\item GndSpd - Ground Speed (GndSpd)
\item Indicated Air Speed (IAS)
\item Lateral Acceleration (LatAc)
\item Normal Acceleration (NormAc)
\item Outside Air Temperature (OAT)
\item Pitch
\item Roll
\item True Airspeed (TAS)
\item Voltage 1 (volt1)
\item Voltage 2 (volt2)
\item Vertical Speed (VSpd)
\item Vertical Speed Gs (VSpdG)
\end{enumerate}
%\end{multicols}

These files had identifying information (fleet identifier, tail number, date and time, and latitude/longitude coordinates) which was removed to protect the identify of the pilots. The data is provided unnormalized.

For this work, two parameters were selected as prediction targets: {\it RPM} and {\it Pitch}. These are interesting as the first (RPM) is a product of engine activity, with other engine related parameters being correlated; while Pitch is most directly influenced by pilot controls, making it particularly challenging to predict.

\subsection{Coal-fired Power Plant Data}
This data set consists of 10 days of per-minute data readings extracted from 12 of the plant's burners. Each of these 12 data files has 12 parameters of time series data:

%\begin{multicols}{2}
\begin{enumerate}
\item Conditioner Inlet Temp
\item Conditioner Outlet Temp
\item Coal Feeder Rate
\item Primary Air Flow
\item Primary Air Split
\item System Secondary Air Flow Total
\item Secondary Air Flow
\item Secondary Air Split
\item Tertiary Air Split
\item Total Combined Air Flow
\item Supplementary Fuel Flow
\item Main Flame Intensity
\end{enumerate}
%\end{multicols}

In order to protect the confidentiality of the power plant which provided the data, along with any sensitive data elements, all identifying data has been scrubbed from the data sets (such as dates, times, locations and facility names). Further, the data has been pre-normalized between 0 and 1 as a further precaution. So while the data cannot be reverse engineered to identify the originating power plant or actual parameter values -- it still is an extremely valuable test data set for times series data prediction as it consists of real world data from a highly complex system with interdependent data streams.

For this data set, two of the parameters were of key interest for time series data prediction, \emph{Main Flame Intensity} and \emph{Supplementary Fuel Flow}. Similar to the choices from the NGAFID data, Main Flame Intensity is mostly a product of conditions within the burner, while Supplementary Fuel Flow is more directly controlled by humans.

\section{Results}
\label{sec:results}

\subsection{Computing Environment} 
%REPLACE UNIVERSITY WITH RIT
Results were gathered using university research computing systems. Compute nodes utilized ranged between 10 core 2.3 GHz Intel\textregistered Xeon\textregistered CPU E5-2650 v3, 32 core 2.6 GHz AMD Opteron\texttrademark Processor 6282 SE and 48 core 2.5 GHz AMD Opteron\texttrademark Processor 6180 SEs, which was unavoidable due to cluster scheduling policies. All compute nodes ran RedHat Enterprise Linux 6.10. This did result in some variation in performance, however discrepancies in timing were overcome by averaging over multiple runs in aggregate. 

\subsection{Experimental Design}
\label{sec:experimental_design}

To better understand how the different memory cells performed in time series data prediction, multiple EXAMM runs were conducted that allowed different types of memory cells. The first set of runs ($5$) only any added cells of only a single, particular memory cell type, \ie, either a $\Delta$-RNN, GRU, LSTM, MGU, or UGRNN. The next set of runs ($5$) was nearly identical, except these allowed nodes to be simple neurons in addition to each particular memory cell type (such runs are appended with a \emph{+simple} in the result tables). One final version was run where all cell types and simple neurons were allowed; resulting in $11$ different EXAMM run types (such runs are labeled as \emph{all} in the result tables). These different types of runs were done for each of the four prediction parameters (RPM, pitch, main flame intensity, and supplementary fuel flow). $K$-fold cross validation was done for each prediction parameter, with a fold size of $2$. This resulted in $5$ folds for the NGAFID data (as it had $10$ flight data files), and $6$ folds for the coal plant data (as it has $12$ burner data files). Each fold and EXAMM setting run was repeated $10$ times.  In total, each of the $11$ EXAMM run types was done $110$ times ($50$ times for the NGAFID data $k$-fold validation and $60$ times for the coal data $k$-fold validation), for a total of $2,420$ separate runs.

% ALEX: I don't think you meant stochastic backprop (which is an algorithm used to train probabilistic neural networks/generative models) --> you meant backprop with SGD, correct? I fixed this b/c stochastic backprop didn't make sense to me...
All neural networks were trained with backpropagation and stochastic gradient descent (SGD) using the same hyperparameters. SGD was run with a learning rate $\eta = 0.001$, utilizing Nesterov momentum with $mu = 0.9$. No dropout regularization was used since it has been shown in other work to reduce performance when training RNNs for time series prediction~\cite{elsaid2018optimizing}. To prevent exploding gradients, gradient clipping (as described by Pascanu \etal~\cite{pascanu2013difficulty}) was used when the norm of the gradient was above a threshold of $1.0$. To improve performance for vanishing gradients, gradient boosting (the opposite of clipping) was used when the norm of the gradient was below a threshold of $0.05$. The forget gate bias of the LSTM cells had 1.0 added to it as this has been shown to yield significant improvements in training time by Jozefowicz \etal~\cite{jozefowicz2015empirical}; otherwise weights were initialized as described in Section~\ref{sec:lamarckian_weight_initialization}.

% ALEX: was mutation actually at 70% ? that's a high mutation rate/prob
%EXAMM hyperparameters
Each EXAMM run consisted of $10$ islands, each with a population size of $5$, and new RNNs were generated via intra-island crossover (at 20\%), mutation (at 70\%), and inter-island crossover at (10\%). All mutation operations (described in Section~\ref{sec:examm}) except for \emph{split edge} were utilized, as \emph{split edge} can be recreated with the \emph{add node} and \emph{disable edge} operations. The $10$ utilized mutation operations were performed each with a uniform 10\% chance.  Each EXAMM run generated $2000$ RNNs, with each RNN being trained for $10$ epochs. These runs were performed utilizing $20$ processors in parallel, and on average required approximately $0.5$ compute hours.  In total, these results come from training over $4,840,000$ RNNs, requiring \textasciitilde $24,200$ CPU hours of compute time.

\subsection{Evolved RNN Performance}

\begin{table*}
\begin{tiny}
\centering
\begin{tabular}{|l|r|r|r|r|r|r|r|r|r|r|r|r|r|r|r|}
\multicolumn{16}{c}{{\bf Flame Intensity}} \\
\hline
  & \multicolumn{3}{|c|}{MSE} & \multicolumn{4}{|c|}{Edges} & \multicolumn{4}{|c|}{Rec. Edges} & \multicolumn{4}{|c|}{Hidden Nodes}\\
\hline
Run Type & Min & Avg & Max  & Min & Avg & Max & Corr. & Min & Avg & Max & Corr. & Min & Avg & Max & Corr.\\
\hline
\hline
all & 0.000438215 & 0.00168176 & 0.00308112 & 11 & 28 & 64 & -0.274 & 0 & 5.7 & 15 & -0.147 & 11 & 16 & 22 & -0.228\\
$\Delta$-RNN & 0.000419515 & \textbf{0.00164756} & 0.00365646 & 15 & 29 & 72 & -0.163 & 1 & 6.3 & 16 & -0.136 & 11 & 16 & 25 & -0.251\\
GRU & 0.000493313 & \textbf{0.00166337} & 0.0035205 & 11 & 28 & 56 & -0.115 & 1 & 6.6 & 20 & -0.101 & 12 & 16 & 21 & 0.109\\
LSTM & 0.000460576 & 0.00174216 & 0.00386718 & 11 & 28 & 53 & -0.187 & 0 & 7.9 & 26 & -0.386 & 10 & 16 & 21 & -0.241\\
MGU & 0.000644687 & 0.00174361 & 0.00342526 & 8 & 27 & 49 & -0.241 & 1 & 6.5 & 19 & -0.127 & 13 & 16 & 21 & -0.373\\
UGRNN & 0.000531725 & 0.00166423 & 0.00399293 & 19 & 28 & 54 & -0.215 & 1 & 6.9 & 25 & -0.192 & 12 & 16 & 22 & -0.209\\
\hline
\multicolumn{16}{c}{ } \\
\multicolumn{16}{c}{{\bf Fuel Flow}} \\
\hline
  & \multicolumn{3}{|c|}{MSE} & \multicolumn{4}{|c|}{Edges} & \multicolumn{4}{|c|}{Rec. Edges} & \multicolumn{4}{|c|}{Hidden Nodes}\\
\hline
Run Type & Min & Avg & Max  & Min & Avg & Max & Corr. & Min & Avg & Max & Corr. & Min & Avg & Max & Corr.\\
\hline
\hline
all & 4.86288e-06 & 0.000134371 & 0.000293377 & 12 & 31 & 55 & 0.138 & 0 & 6.7 & 18 & 0.15 & 8 & 16 & 22 & 0.251\\
$\Delta$-RNN & 6.44047e-06 & 0.000140518 & 0.000306163 & 15 & 29 & 57 & -0.0995 & 0 & 6.8 & 17 & -0.264 & 12 & 16 & 22 & -0.194\\
GRU & 1.65072e-05 & \textbf{0.000131481} & 0.000289922 & 16 & 31 & 66 & -0.193 & 0 & 6.6 & 29 & -0.209 & 10 & 16 & 22 & -0.146\\
LSTM & 5.90673e-06 & \textbf{0.000121158} & 0.000262113 & 14 & 31 & 65 & -0.0119 & 1 & 6.8 & 22 & -0.028 & 11 & 16 & 23 & 0.0572\\
MGU & 1.83369e-05 & 0.00013616 & 0.000281395 & 13 & 32 & 63 & -0.235 & 0 & 6.7 & 17 & -0.303 & 7 & 16 & 23 & -0.141\\
UGRNN & 6.08753e-06 & 0.000143734 & 0.000341757 & 10 & 29 & 78 & 0.0308 & 0 & 6.7 & 19 & -0.0971 & 9 & 16 & 22 & 0.158\\
\hline
\multicolumn{16}{c}{ } \\
\multicolumn{16}{c}{{\bf RPM}} \\
\hline
  & \multicolumn{3}{|c|}{MSE} & \multicolumn{4}{|c|}{Edges} & \multicolumn{4}{|c|}{Rec. Edges} & \multicolumn{4}{|c|}{Hidden Nodes}\\
\hline
Run Type & Min & Avg & Max  & Min & Avg & Max & Corr. & Min & Avg & Max & Corr. & Min & Avg & Max & Corr.\\
\hline
\hline
all & 0.00410034 & 0.00725003 & 0.0142856 & 25 & 36 & 54 & 0.215 & 0 & 5.3 & 13 & -0.133 & 24 & 29 & 33 & 0.0921\\
$\Delta$-RNN & 0.00247046 & \textbf{0.0066479} & 0.0132178 & 23 & 37 & 66 & 0.00237 & 0 & 4.8 & 16 & -0.083 & 25 & 29 & 35 & -0.00169\\
GRU & 0.00221411 & \textbf{0.00684006} & 0.0108375 & 24 & 36 & 72 & -0.207 & 0 & 5.9 & 16 & -0.342 & 25 & 29 & 36 & -0.262\\
LSTM & 0.00323368 & 0.00736006 & 0.014325 & 24 & 35 & 46 & 0.018 & 0 & 4.7 & 14 & -0.437 & 23 & 28 & 30 & -0.14\\
MGU & 0.00343788 & 0.00730802 & 0.0153965 & 28 & 38 & 57 & 0.057 & 0 & 5.3 & 14 & -0.519 & 25 & 29 & 33 & 0.124\\
UGRNN & 0.00307084 & 0.00716817 & 0.0116206 & 25 & 36 & 66 & -0.277 & 0 & 5.3 & 15 & -0.373 & 25 & 29 & 36 & -0.315\\
\hline
\multicolumn{16}{c}{ } \\
\multicolumn{16}{c}{{\bf Pitch}} \\
\hline
  & \multicolumn{3}{|c|}{MSE} & \multicolumn{4}{|c|}{Edges} & \multicolumn{4}{|c|}{Rec. Edges} & \multicolumn{4}{|c|}{Hidden Nodes}\\
\hline
Run Type & Min & Avg & Max  & Min & Avg & Max & Corr. & Min & Avg & Max & Corr. & Min & Avg & Max & Corr.\\
\hline
\hline
all & 0.00101445 & 0.00347982 & 0.00582811 & 22 & 35 & 67 & -0.132 & 0 & 3 & 10 & -0.223 & 24 & 28 & 35 & -0.109\\
$\Delta$-RNN & 0.00149607 & 0.00328248 & 0.00557884 & 24 & 35 & 49 & 0.0524 & 0 & 3.4 & 12 & -0.114 & 23 & 28 & 32 & 0.104\\
GRU & 0.00133482 & \textbf{0.00323731} & 0.00506138 & 22 & 34 & 54 & -0.0691 & 0 & 3 & 11 & -0.212 & 24 & 28 & 32 & -0.113\\
LSTM & 0.00114187 & 0.0033505 & 0.00558931 & 26 & 35 & 54 & -0.147 & 0 & 3.5 & 9 & -0.126 & 24 & 28 & 33 & -0.117\\
MGU & 0.00148457 & \textbf{0.00327753} & 0.00565053 & 26 & 35 & 51 & 0.0707 & 0 & 3.3 & 11 & -0.388 & 25 & 28 & 32 & 0.0442\\
UGRNN & 0.00117033 & 0.0033487 & 0.00559272 & 22 & 33 & 50 & -0.0981 & 0 & 3 & 13 & -0.165 & 23 & 28 & 32 & -0.0133\\
\hline
\end{tabular}

\caption{\label{table:solo_all_stats} Statistics for RNNs Evolved With Individual Memory Cells and RNNs Evolved With All Memory Types.}
\end{tiny}
\vspace{-5mm}
\end{table*}

\begin{table*}
\begin{tiny}
\centering
\begin{tabular}{|l|r|r|r|r|r|r|r|r|r|r|r|r|r|r|r|r|r|r|r|}
\multicolumn{20}{c}{{\bf Flame Intensity}} \\
\hline
  & \multicolumn{3}{|c|}{MSE} & \multicolumn{4}{|c|}{Edges} & \multicolumn{4}{|c|}{Rec. Edges} & \multicolumn{4}{|c|}{Memory Cells} & \multicolumn{4}{|c|}{Simple Neurons}\\
\hline
Run Type & Min & Avg & Max  & Min & Avg & Max & Corr. & Min & Avg & Max & Corr. & Min & Avg & Max & Corr. & Min & Avg & Max & Corr.\\
\hline
\hline
$\Delta$-RNN+simple & 0.00042112 & \textbf{0.0015147} & 0.0038971 & 12 & 27 & 62 & -0.25 & 1 & 7.7 & 20 & -0.222 & 0 & 1.9 & 7 & -0.137 & 0 & 1.6 & 6 & -0.238\\
GRU+simple & 0.00068848 & 0.0016726 & 0.0039989 & 18 & 29 & 55 & -0.23 & 0 & 7.1 & 17 & -0.11 & 0 & 1.8 & 4 & -0.133 & 0 & 1.9 & 6 & -0.384\\
LSTM+simple & 0.00046443 & \textbf{0.0015194} & 0.0032395 & 19 & 31 & 65 & -0.21 & 1 & 7.3 & 18 & -0.208 & 0 & 1.6 & 6 & -0.27 & 0 & 2.2 & 7 & 0.0671\\
MGU+simple & 0.00053109 & 0.0016088 & 0.003241 & 18 & 29 & 55 & -0.19 & 0 & 7.8 & 21 & -0.238 & 0 & 1.6 & 6 & -0.00329 & 0 & 2 & 7 & -0.213\\
UGRNN+simple & 0.00043835 & 0.0016874 & 0.0041592 & 16 & 29 & 50 & -0.31 & 1 & 6.2 & 16 & -0.274 & 0 & 1.5 & 5 & -0.349 & 0 & 2.4 & 6 & -0.061\\
\hline
\multicolumn{20}{c}{ } \\
\multicolumn{20}{c}{{\bf Fuel Flow}} \\
\hline
  & \multicolumn{3}{|c|}{MSE} & \multicolumn{4}{|c|}{Edges} & \multicolumn{4}{|c|}{Rec. Edges} & \multicolumn{4}{|c|}{Memory Cells} & \multicolumn{4}{|c|}{Simple Neurons}\\
\hline
Run Type & Min & Avg & Max  & Min & Avg & Max & Corr. & Min & Avg & Max & Corr. & Min & Avg & Max & Corr. & Min & Avg & Max & Corr.\\
\hline
\hline
$\Delta$-RNN+simple & 8.1558e-06 & \textbf{0.00012294} & 0.00026373 & 9 & 29 & 55 & -0.12 & 0 & 6.4 & 15 & -0.026 & 0 & 1.7 & 5 & -0.2 & 0 & 1.8 & 6 & -0.00871\\
GRU+simple & 1.3193e-05 & 0.00013061 & 0.00028389 & 9 & 31 & 62 & -0.17 & 0 & 6 & 21 & 0.00366 & 0 & 2.1 & 6 & -0.196 & 0 & 1.8 & 6 & 0.0181\\
LSTM+simple & 8.6606e-06 & 0.00013071 & 0.00025847 & 14 & 31 & 61 & -0.098 & 0 & 6.5 & 22 & -0.187 & 0 & 2.2 & 7 & -0.0819 & 0 & 1.9 & 5 & 0.0817\\
MGU+simple & 7.2707e-06 & \textbf{0.00012367} & 0.00028356 & 11 & 33 & 61 & -0.072 & 1 & 6.8 & 26 & -0.228 & 0 & 2.2 & 7 & -0.107 & 0 & 2.1 & 8 & -0.0846\\
UGRNN+simple & 5.6189e-06 & 0.00014337 & 0.00030186 & 12 & 32 & 62 & -0.15 & 0 & 6.1 & 15 & -0.253 & 0 & 1.8 & 6 & -0.0961 & 0 & 2.3 & 8 & -0.102\\
\hline
\multicolumn{20}{c}{ } \\
\multicolumn{20}{c}{{\bf RPM}} \\
\hline
  & \multicolumn{3}{|c|}{MSE} & \multicolumn{4}{|c|}{Edges} & \multicolumn{4}{|c|}{Rec. Edges} & \multicolumn{4}{|c|}{Memory Cells} & \multicolumn{4}{|c|}{Simple Neurons}\\
\hline
Run Type & Min & Avg & Max  & Min & Avg & Max & Corr. & Min & Avg & Max & Corr. & Min & Avg & Max & Corr. & Min & Avg & Max & Corr.\\
\hline
\hline
$\Delta$-RNN+simple & 0.0029348 & 0.0069672 & 0.011182 & 25 & 36 & 48 & -0.17 & 0 & 4.8 & 14 & -0.496 & 0 & 0.88 & 3 & 0.0549 & 0 & 1.2 & 4 & -0.142\\
GRU+simple & 0.0039148 & 0.0070081 & 0.012816 & 29 & 38 & 63 & -0.054 & 1 & 5.1 & 13 & -0.049 & 0 & 1.2 & 4 & -0.0963 & 0 & 1.3 & 5 & 0.0855\\
LSTM+simple & 0.0027877 & \textbf{0.0063965} & 0.01091 & 27 & 37 & 60 & -0.11 & 0 & 4.6 & 11 & -0.5 & 0 & 1 & 3 & -0.223 & 0 & 1.2 & 4 & 0.0936\\
MGU+simple & 0.0024288 & \textbf{0.0065725} & 0.011325 & 24 & 36 & 44 & 0.051 & 0 & 5.6 & 13 & -0.391 & 0 & 0.98 & 3 & -0.0944 & 0 & 1.1 & 3 & 0.178\\
UGRNN+simple & 0.0037114 & 0.0073811 & 0.012252 & 22 & 38 & 48 & -0.11 & 0 & 4.3 & 10 & -0.304 & 0 & 1.1 & 4 & -0.00488 & 0 & 1.4 & 4 & -0.0642\\
\hline
\multicolumn{20}{c}{ } \\
\multicolumn{20}{c}{{\bf Pitch}} \\
\hline
  & \multicolumn{3}{|c|}{MSE} & \multicolumn{4}{|c|}{Edges} & \multicolumn{4}{|c|}{Rec. Edges} & \multicolumn{4}{|c|}{Memory Cells} & \multicolumn{4}{|c|}{Simple Neurons}\\
\hline
Run Type & Min & Avg & Max  & Min & Avg & Max & Corr. & Min & Avg & Max & Corr. & Min & Avg & Max & Corr. & Min & Avg & Max & Corr.\\
\hline
\hline
$\Delta$-RNN+simple & 0.0015138 & 0.0032316 & 0.0053377 & 20 & 34 & 54 & -0.14 & 0 & 2.6 & 8 & -0.341 & 0 & 0.86 & 4 & -0.00809 & 0 & 1.1 & 5 & -0.168\\
GRU+simple & 0.0011566 & \textbf{0.0032083} & 0.0067072 & 22 & 36 & 57 & 0.056 & 0 & 3.2 & 9 & 0.0472 & 0 & 0.98 & 4 & 0.042 & 0 & 1 & 3 & -0.00732\\
LSTM+simple & 0.0010335 & 0.0032365 & 0.005414 & 22 & 34 & 52 & -0.25 & 0 & 2.9 & 10 & -0.233 & 0 & 0.9 & 3 & -0.176 & 0 & 0.98 & 4 & -0.157\\
MGU+simple & 0.0010133 & 0.0033326 & 0.0060865 & 25 & 35 & 53 & -0.15 & 0 & 2.8 & 11 & -0.307 & 0 & 0.88 & 3 & -0.276 & 0 & 0.98 & 5 & 0.163\\
UGRNN+simple & 0.0015551 & \textbf{0.0031352} & 0.0053126 & 22 & 34 & 52 & 0.1 & 0 & 3.1 & 8 & -0.216 & 0 & 0.94 & 3 & -0.063 & 0 & 1 & 4 & 0.181\\
\hline
\end{tabular}

\caption{\label{table:with_ff_stats} Statistics for RNNs Evolved With Simple Neurons and Memory Cells.}
\end{tiny}
\vspace{-5mm}
\end{table*}

\begin{table*}
\begin{tiny}
\centering
\begin{tabular}{|l|r|r|r|r|r|r|r|r|r|r|r|r|r|r|r|r|r|r|r|r|r|r|r|r|r|r|r|r|}
\hline
 & \multicolumn{4}{|c|}{Simple}& \multicolumn{4}{|c|}{LSTM}& \multicolumn{4}{|c|}{UGRNN}& \multicolumn{4}{|c|}{$\Delta$-RNN}& \multicolumn{4}{|c|}{MGU}& \multicolumn{4}{|c|}{GRU}\\
\hline
Run Type & Min & Avg & Max & Corr & Min & Avg & Max & Corr & Min & Avg & Max & Corr & Min & Avg & Max & Corr & Min & Avg & Max & Corr & Min & Avg & Max & Corr\\
\hline
\hline
flame & 0 & 0.6 & 3&0.044 & 0 & 0.5 & 3&-0.26 & 0 & 0.5 & 4&-0.11 & 0 & 0.5 & 2&-0.12 & 0 & 0.5 & 3&0.073 & 0 & 0.7 & 4&-0.17\\
fuel flow & 0 & 0.8 & 3&0.0076 & 0 & 0.6 & 4&0.11 & 0 & 0.7 & 4&-0.061 & 0 & 0.6 & 4&-0.041 & 0 & 0.6 & 2&0.092 & 0 & 0.8 & 3&0.018\\
pitch & 0 & 0.3 & 3&-0.017 & 0 & 0.3 & 3&-0.031 & 0 & 0.3 & 2&0.032 & 0 & 0.2 & 2&0.052 & 0 & 0.5 & 2&-0.036 & 0 & 0.3 & 2&-0.32\\
rpm & 0 & 0.4 & 2&-0.14 & 0 & 0.4 & 2&0.24 & 0 & 0.3 & 2&-0.086 & 0 & 0.4 & 2&0.026 & 0 & 0.4 & 3&0.32 & 0 & 0.2 & 2&-0.16\\
\hline
\end{tabular}
\caption{\label{table:all_statistics} Hidden Node Counts and Correlations to Mean Square Error for EXAMM Runs Using All Cell Types.}
\end{tiny}
\vspace{-5mm}
\end{table*}

\begin{table}
\begin{tiny}
\centering
\begin{tabular}{|r|l||r|l||r|l|}
\multicolumn{6}{c}{{\bf Flame Intensity}} \\
\hline
\multicolumn{2}{|c||}{{\bf Best Case}} & \multicolumn{2}{|c||}{{\bf Avg. Case}} & \multicolumn{2}{|c|}{{\bf Worst Case}} \\
\hline
          $\Delta$-RNN &        -0.92312 &        $\Delta$-RNN+simple &         -1.7775 &             all &         -1.5404\\
       $\Delta$-RNN+simple &        -0.90534 &         LSTM+simple &         -1.7148 &         LSTM+simple &         -1.1066\\
            all &        -0.71602 &          MGU+simple &        -0.53749 &          MGU+simple &         -1.1026\\
       UGRNN+simple &        -0.71451 &           $\Delta$-RNN &       -0.026901 &             MGU &        -0.59787\\
           LSTM &        -0.46836 &             GRU &         0.18143 &             GRU &        -0.33703\\
        LSTM+simple &        -0.42565 &           UGRNN &         0.19272 &           $\Delta$-RNN &        0.035348\\
            GRU &        -0.10578 &          GRU+simple &         0.30281 &            LSTM &         0.61246\\
         MGU+simple &         0.31264 &             all &         0.42371 &        delta+simple &         0.69439\\
          UGRNN &         0.31964 &        UGRNN+simple &         0.49785 &           UGRNN &          0.9569\\
            MGU &          1.5708 &            LSTM &          1.2196 &          GRU+simple &         0.97318\\
         GRU+simple &          2.0557 &             MGU &          1.2386 &        UGRNN+simple &          1.4123\\
\hline
\multicolumn{6}{c}{ } \\
\multicolumn{6}{c}{{\bf Fuel flow}} \\
\hline
\multicolumn{2}{|c||}{{\bf Best Case}} & \multicolumn{2}{|c||}{{\bf Avg. Case}} & \multicolumn{2}{|c|}{{\bf Worst Case}} \\
\hline
            all &        -0.92643 &            LSTM &         -1.4415 &         LSTM+simple &         -1.2349\\
       UGRNN+simple &         -0.7644 &        $\Delta$-RNN+simple &         -1.2172 &            LSTM &         -1.0818\\
           LSTM &        -0.70271 &          MGU+simple &         -1.1255 &        $\Delta$-RNN+simple &          -1.014\\
          UGRNN &        -0.66396 &          GRU+simple &        -0.25195 &             MGU &        -0.27097\\
          $\Delta$-RNN &        -0.58832 &         LSTM+simple &        -0.23921 &          MGU+simple &         -0.1799\\
         MGU+simple &        -0.41037 &             GRU &        -0.14222 &          GRU+simple &        -0.16598\\
       $\Delta$-RNN+simple &        -0.22068 &             all &         0.22163 &             GRU &        0.087564\\
        LSTM+simple &         -0.1125 &             MGU &         0.44679 &             all &         0.23284\\
         GRU+simple &         0.85882 &           $\Delta$-RNN &         0.99531 &        UGRNN+simple &         0.58938\\
            GRU &          1.5692 &        UGRNN+simple &          1.3537 &           $\Delta$-RNN &         0.77052\\
            MGU &          1.9613 &           UGRNN &          1.4002 &           UGRNN &          2.2672\\
\hline
\multicolumn{6}{c}{ } \\
\multicolumn{6}{c}{{\bf RPM}} \\
\hline
\multicolumn{2}{|c||}{{\bf Best Case}} & \multicolumn{2}{|c||}{{\bf Avg. Case}} & \multicolumn{2}{|c|}{{\bf Worst Case}} \\
\hline
            GRU &          -1.444 &         LSTM+simple &         -1.7472 &             GRU &         -1.0958\\
         MGU+simple &         -1.1012 &          MGU+simple &         -1.2299 &         LSTM+simple &         -1.0499\\
          $\Delta$-RNN &         -1.0347 &           $\Delta$-RNN &         -1.0081 &        $\Delta$-RNN+simple &        -0.87687\\
        LSTM+simple &        -0.52825 &             GRU &         -0.4433 &          MGU+simple &        -0.78566\\
       $\Delta$-RNN+simple &        -0.29348 &        $\Delta$-RNN+simple &       -0.069508 &           UGRNN &        -0.59783\\
          UGRNN &       -0.076276 &          GRU+simple &        0.050686 &        UGRNN+simple &        -0.19645\\
           LSTM &         0.18368 &           UGRNN &         0.52115 &          GRU+simple &         0.16258\\
            MGU &         0.50967 &             all &         0.76179 &           $\Delta$-RNN &         0.41787\\
       UGRNN+simple &          0.9463 &             MGU &         0.93224 &             all &          1.0968\\
         GRU+simple &           1.271 &            LSTM &          1.0852 &            LSTM &          1.1219\\
            all &          1.5672 &        UGRNN+simple &           1.147 &             MGU &          1.8033\\
\hline
\multicolumn{6}{c}{ } \\
\multicolumn{6}{c}{{\bf Pitch}} \\
\hline
\multicolumn{2}{|c||}{{\bf Best Case}} & \multicolumn{2}{|c||}{{\bf Avg. Case}} & \multicolumn{2}{|c|}{{\bf Worst Case}} \\
\hline
         MGU+simple &         -1.1631 &        UGRNN+simple &         -1.6163 &             GRU &         -1.3295\\
            all &         -1.1577 &          GRU+simple &        -0.82052 &        UGRNN+simple &        -0.76284\\
        LSTM+simple &         -1.0698 &        $\Delta$-RNN+simple &        -0.56665 &        $\Delta$-RNN+simple &        -0.70622\\
           LSTM &         -0.5688 &         LSTM+simple &        -0.51389 &         LSTM+simple &        -0.53415\\
         GRU+simple &        -0.50079 &             GRU &         -0.5047 &           $\Delta$-RNN &        -0.16235\\
          UGRNN &        -0.43726 &             MGU &       -0.066984 &            LSTM &        -0.13873\\
            GRU &         0.32298 &           delta &       -0.013118 &           UGRNN &        -0.13104\\
            MGU &          1.0151 &          MGU+simple &         0.53287 &             MGU &     -0.00065639\\
          $\Delta$-RNN &          1.0682 &           UGRNN &         0.70761 &             all &         0.39991\\
       $\Delta$-RNN+simple &          1.1501 &            LSTM &         0.72719 &          MGU+simple &         0.98284\\
       UGRNN+simple &          1.3411 &             all &          2.1345 &          GRU+simple &          2.3828\\
\hline
\multicolumn{6}{c}{ } \\
\multicolumn{6}{c}{{\bf Overall Combined}} \\
\hline
\multicolumn{2}{|c||}{{\bf Best Case}} & \multicolumn{2}{|c||}{{\bf Avg. Case}} & \multicolumn{2}{|c|}{{\bf Worst Case}} \\
\hline
         MGU+simple &        -0.59051 &         LSTM+simple &         -1.0538 &         LSTM+simple &        -0.98141\\
        LSTM+simple &        -0.53405 &        $\Delta$-RNN+simple &        -0.90771 &             GRU &         -0.6687\\
           LSTM &        -0.38905 &          MGU+simple &        -0.59001 &        $\Delta$-RNN+simple &        -0.47566\\
          $\Delta$-RNN &        -0.36948 &             GRU &         -0.2272 &          MGU+simple &        -0.27133\\
            all &        -0.30824 &          GRU+simple &        -0.17974 &             all &        0.047292\\
          UGRNN &        -0.21446 &           $\Delta$-RNN &       -0.013211 &            LSTM &         0.12847\\
       $\Delta$-RNN+simple &       -0.067358 &        UGRNN+simple &         0.34556 &             MGU &         0.23345\\
            GRU &        0.085614 &            LSTM &         0.39761 &        UGRNN+simple &         0.26059\\
       UGRNN+simple &         0.20212 &             MGU &         0.63765 &           $\Delta$-RNN &         0.26535\\
         GRU+simple &          0.9212 &           UGRNN &         0.70542 &           UGRNN &         0.62381\\
            MGU &          1.2642 &             all &         0.88541 &          GRU+simple &         0.83814\\
\hline
\end{tabular}
\caption{\label{table:all_rankings} EXAMM Run Types Prediction Error Ranked By Standard Deviation From Mean.}
\end{tiny}
\vspace{-10mm}
\end{table}

% Central results/synthesis
Table~\ref{table:solo_all_stats} shows aggregated results for each of the $50$ or $60$ EXAMM runs done ($5$ or $6$ folds, each with $10$ repeats) that allowed only one of each memory cell type, along with the EXAMM runs that allowed for added nodes to be of all node types (\ie, \emph{all}). Table~\ref{table:with_ff_stats} further shows the aggregated results for runs which allowed both simple neurons and one particular cell type (\ie, \emph{+simple}).

These tables present the minimum, average, and maximum mean squared error (MSE) on average across the runs. Table~\ref{table:solo_all_stats} also shows the minimum, average, and maximum number of hidden nodes, which would be entirely of the one memory cell type, or of any memory cell type or simple neurons in the case of the \emph{all} runs, as well as how the the number of nodes correlate to the MSE. Similar statistics are shown for the numbers of feed forward edges and the numbers of recurrent edges. Note that as a lower MSE is better, a negative correlation means that having more nodes or edges was correlated to lower MSE.  Table~\ref{table:with_ff_stats} also divides hidden nodes into counts for simple neurons and number of memory cell nodes for a certain run type. Top $2$ best models (avg) scores are shown in \textbf{bold}.

The best found MSE scores across the four prediction parameters had a wide range, which could be expected due to varying complexities. As such, to properly rank performance of different run types a metric was needed. Table~\ref{table:all_rankings} orders the $11$ different run types by how many standard deviations the results were from the mean for each prediction parameter. It also provides combined rankings, averaging the deviation from the mean across the four prediction parameters. Each of these tables are ordered from best to worst -- a negative deviation from the mean is that many standard deviations less than the average MSE, and lower MSE is better.%best performing to worst

The results of these experiments led to some interesting findings which the authors feel can help inform further development of neuro-evolution algorithms as well as RNN memory cells. Many of these findings can also serve as warnings to those looking to train well performing RNNs for time series prediction. We summarize the main takeaways from these results as follows:

\noindent
\paragraph{No memory structure was truly the best:} In the overall rankings (Table~\ref{table:all_rankings}), the $\Delta$-RNN, LSTM, and MGU cells seemed to have the highest rankings for best overall performance as well as average and worst case performance, with GRU cells being slightly behind. While UGRNN nodes did not do so well in the total ranking, it should be noted that when coupled with simple neurons, they did perform 2nd best for the best case for fuel flow, and were the best for the average case in predicting pitch. This highlights the importance of testing a wide selection of memory cell types when developing RNNs, as there is \emph{no free lunch} in machine learning -- each memory cell type had its own strengths and weaknesses. It is valuable to note that one of the simplest memory cell types, \ie, the $\Delta$-RNN, performs consistently and competitively with the more complicated, multi-gate LSTM (at the top of the rankings), which is consistent with a growing body of results \cite{ororbia2017diff,ororbia2018using,ororbia2018learned}.

\noindent
\paragraph{Adding simple neurons generally helped - with some notable exceptions:} When looking at the overall rankings, when simple neurons were added as an option to the neuro-evolution process, the networks performed better.  The only exception to this was GRU cells, which tended to perform worse when simple neurons were allowed.  These results may indicate that these memory cells lack the capability to track some kinds of dependencies which the additional simple neurons make up for; this means that there is potentially room to improve these cell structures to capture whatever the simple neurons were providing.  Further examination of why the GRU cells performed worse with simple neurons compared to the other memory cells may help determine the cause of this and makes for an interesting direction for future RNN work.

Another very interesting finding was that utilizing simple neurons with MGU cells resulted in a dramatic improvement, bringing them from some of the worst rankings to some of the best rankings (\eg, in the overall rankings for best found networks, MGU cells alone performed the worst while MGU and simple neurons performed the best). Other cell types (LSTM and $\Delta$-RNN) showed less of an improvement. This finding may highlight that the MGU cells could stand to benefit from further development. This should serve as a warning to others developing neuro-evolution algorithms, in that even the rather simple change of allowing simple neurons can result in significant changes in RNN predictive ability. Selection of node and cell types for neuro-evolution should be done carefully.

\noindent
\paragraph{Allowing all memory cells has risks and benefits:} The authors had hoped that allowing the neuro-evolution process to simply select from all of the memory cell types would allow it to find and utilize whichever cells were most suited to the prediction problem.  Unfortunately, this was not always the case.  While using all memory cell types generally performed better than the mean on the best case, in the average and worst cases, it performed worse than the mean. This was most likely due to the fact that whenever a node was added to the network it could have been from any of the 6 types, so choosing from them uniformly at random ended up sometimes selecting the nodes not best to the task (as they improved the population, but not as well as another memory cell choice could have done). These results are backed up by Table~\ref{table:all_statistics}, which shows the correlations between node types and MSE (again, a negative correlation means more of that cell type resulted in a lower/better MSE), as well as the min, average and max memory cell counts among the networks found by EXAMM. 

That being said, allowing all cell types did find the best networks in the case of fuel flow, 2nd best in the case of pitch, and 3rd best in the case of flame intensity; which is impressive given the larger search space for the neuro-evolution process. We expect to further improve this result by dynamically adapting the rates at which memory cells are generated in part based on how well they have improved the population in the past (with caveats described in the next point) -- this stands as future work which can make EXAMM an even stronger option for generating RNNs, especially in the average and worse cases where they did not fare as well.

\noindent
\paragraph{Larger networks tended to perform better, yet memory cell count correlation to MSE was not a great indicator of which cells performed the best:}  This last point raises some significant challenges for developing neuro-evolution algorithms. When looking at Table~\ref{table:all_statistics} and examining the memory cells types most correlated to improved performance against the memory cell types most frequently selected by EXAMM, meant that EXAMM was not selecting cell types that would produce the best performing RNNs. This may due to the fact that, in some cases, an RNN with a small number of well trained memory cells was sufficient to yield good predictions, and adding more cells to the network only served to confuse the predictions.

The implications of this are two fold: 1), running a neuro-evolution strategy allowing all memory cell types and then utilizing counts or correlations to select a single memory cell type for future runs may not produce the best results, and 2), dynamically tuning which memory cells are selected by a neuro-evolution strategy is more challenging since the process may not select the best cell types (\eg, when the network already has enough memory cells) -- so this would at least need to be coupled with another strategy to determine when the network is ``big enough''.

\section{Conclusions and Future Work}
\label{sec:conclusion}

This work introduced a new neuro-evolution algorithm, Evolutionary eXploration of Augmenting Memory Models (EXAMM), for evolving recurrent neural architectures by directly incorporating powerful memory cells such as the $\Delta$-RNN, MGU, GRU, LSTM and UGRNN units into the evolutionary process. EXAMM was evaluated on the task of predicting $4$ different parameters from two large, real world time-series datasets.  By using repeated $k$-fold cross validation and high performance computing, enough RNNs were evolved to be rigorously analyzed -- a methodology the authors think should be highlighted as novel.  Instead of utilizing them to outperform other algorithms on benchmarks, neuro-evolutionary processes can be used as selection methodologies, providing deeper insights into what neural structures perform best on certain tasks.

Key findings from this work show that a neuro-evolution strategy that selects from a wide number of memory cell structures can yield performant architectures. However, it does so at the expense of reliability in the average and worst cases.  Furthermore, a simple modification to the evolutionary process, \ie, allowing simple neurons, can have dramatic effects on network performance.  In general, while this largely benefits most memory cells, outlier cases showed wide swings from worst to best and best to worst performance. The authors hope that these results will guide future memory cell development, as the addition of simple neurons dramatically improved MGU performance, but also decreased GRU performance. Understanding cases like that of the GRU could yield improvements in cell design. Results showed that cell selection does not necessarily correlate well to the best cell types for a particular problem, partly due to the fact that good cells may not necessarily require a large network. These results should serve as cautionary information for future development of neuro-evolution algorithms.%self-optimizing <-- Alex: cut this, since I don't believe EXAMM tunes its own hyper-parameters, right?

This paper opens up many avenues of future work.  This includes extending EXAMM's search process to allow cellular elements of the underlying neural model to be evolved (as in Rawal and Miikulainen~\cite{rawal-evolving-rnns-2018}); evolving over a large set of post-activation functions; allowing for stochastic operations, \eg, Bernoulli sampling; and incorporating operators such as convolution (to handle video sequences/time series) or its simple approximation, the perturbative operator \cite{juefei2018perturbative}.  Additional future work will involve various hyperparameter optimization strategies to dynamically determine RNN training metaparameters as well as what probabilities EXAMM uses to choose memory cell structures and what probabilities it uses for the mutation and recombination operators. Lastly, implementing even larger scale mutation operations, such as multi-node/layer mutations could potentially speed up EXAMM's neuro-evolution process even further.

%DONT FORGET TO ACKNOWLEDGE MICROBEAM AND NGAFID FOR DATA
%\input{10-acklowedgements}

%\newpage
%\input{08-appendix}

\newpage
\bibliographystyle{IEEEtran}
\bibliography{references}

\end{document}